# Agentic AI for Substance Use Education: Integrating Regulatory and Scientific Knowledge Sources

Kosar Haghani1* · Zahra Kolagar2 · Mohammed Atiquzzaman3

1 Department of Social Science, Texas Woman's University, Denton, Texas, USA. 0000-0002-1129-664X
2 Department of Computer Science, Augsburg University of Applied Sciences, Augsburg, Germany
3 School of Computer Science, University of Oklahoma, Norman, Oklahoma, USA

*Corresponding author: Kosar Haghani, Email: Khaghani@twu.edu

## ABSTRACT

The delivery of traditional substance education has remained problematic due to challenges in scalability, personalization, and the currency of information in a rapidly evolving substance use landscape. While artificial intelligence (AI) offers a promising frontier for enhancing educational delivery, its application in providing real-time, authoritative substance use education remains largely underexplored. We built an agentic-based AI web application that combined Drug Enforcement Administration [1] records with peer-reviewed literature in real-time to provide transparent context-sensitive substance use education. The system uses retrieval-augmented generation with a carefully filtered corpus of 102 documents and dynamic PubMed [2,3] queries. Document storage was semantically chunked and placed in a vector representation in order to be easily retrieved. We conducted an expert evaluation study in which a panel of five subject matter experts generated 30 domain-specific questions, and two independent raters assessed 90 system interactions (30 primary questions plus two contextual follow-ups each) using a five-point Likert scale across four criteria: factual accuracy, citation quality, contextual coherence, and regulatory appropriateness. Mean ratings ranged from 4.18 to 4.35 across the four criteria (overall category range: 4.05–4.52), with substantial inter-rater agreement (Cohen's $\kappa=0.78$). These findings suggest that agentic AI architectures integrating authoritative regulatory sources with real-time scientific literature represent a promising direction for scalable, accurate, and verifiable health education delivery, warranting further evaluation through longitudinal user studies.

**Keywords:** Agentic AI, Substance Use Education, Retrieval-Augmented Generation, Public Health

## INTRODUCTION

Substance use and addiction affect more than 35 million people worldwide with significant social, economic, and health consequences [4]. Despite decades of investment in prevention and education programs, severe limitations persist in engagement, personalization, and scalability [5-7]. Information-seeking behaviors have changed radically, with people increasingly turning to digital platforms for instant health advice, creating a discrepancy between traditional educational delivery and user needs.

The application of general-purpose large language models (LLMs) in substance use education is hindered by critical limitations that compromise user safety. Despite their conversational fluency, these models suffer from "medical hallucinations" generating plausible but factually incorrect clinical details, such as inaccurate drug dosages or outdated safety guidelines [8]. Standard LLMs rely on fixed training cutoffs, making them incapable of reflecting real-time volatility in the substance use landscape, such as emerging synthetic opioids or shifting regulatory updates [9,10]. Recent studies reveal significant risks of 'algorithmic stigma,' where models mirror and amplify societal biases, using stigmatizing language that can alienate individuals seeking help for substance use disorders [11,12]. This concern is particularly acute given the well-documented comorbidity between substance use and trauma-related conditions such as post-traumatic stress disorder, especially among adolescents [13], who represent a vulnerable population requiring trauma-informed approaches in any AI-mediated educational intervention. The "black box" nature presents a transparency crisis; without retrieval-augmented generation (RAG) providing verifiable source attribution, users and educators cannot audit AI advice, breaking the evidence-based "chain of trust" required for health interventions [14,15].

Retrieval-augmented generation (RAG) and agentic AI architectures offer solutions to these limitations. Unlike passive language models, agentic systems are dynamic, autonomous agents that reason, plan, and interact with external knowledge sources [16,17]. RAG combines pre-trained parametric memory with non-parametric memory (dense vector indices of documents) for knowledge-intensive tasks, enabling models to generate more specific, diverse, and factual language than parametric-only baselines [18]. They maintain task focus, retrieve information from external resources, synthesize diverse information, and provide evidence-based answers [19]. For substance use education, these abilities contextualize information within regulatory frameworks and scientific findings, addressing key gaps in existing educational technologies.

Conversational AI has emerged as a promising modality for delivering substance use interventions, offering accessibility, scalability, and reduced stigma compared to traditional services [20-22]. Lee et al. [15] systematically reviewed 28 studies of chatbot-assisted interventions for substance use, finding that while these interventions demonstrated promising effectiveness across alcohol, smoking, and drug use applications, most operated as standalone systems without integrated authoritative knowledge sources, and the authors emphasized the need for ongoing monitoring of feedback and guidance to ensure safe intervention delivery. Ogilvie et al. [23] emphasized that chatbots remain underutilized in mental health services despite emerging evidence supporting their effectiveness. Medical AI research has shown that LLMs demonstrate capabilities in answering medical questions and summarizing complex information, but their application in clinical contexts demands attention to accuracy, explainability, and integration with evidence-based guidelines [17,24,25].

However, agentic systems for substance use education remain understudied. Current applications have been conducted in isolated chatbot implementations without formal incorporation of authoritative policy sources or real-time scientific literature [5,26]. Psychosocial risk factors, including family conflict, marital dissolution, and gender-based stressors, are not usually addressed, though they correlate with substance use habits [26,27]. Ethical issues regarding data governance, algorithmic bias, and cultural sensitivity require transparent, accountable, evidence-based system design [28,29]. Evaluation frameworks emphasize proximal outcomes such as engagement and user satisfaction while neglecting systematic assessment of educational effectiveness and information accuracy. The field lacks demonstrated examples of agentic AI architectures deployed for educational purposes with rigorous evaluation [30-37].

This study addresses these gaps by building and testing an agentic RAG system for substance use education. Three cornerstone elements were incorporated: (1) official Drug Enforcement Administration (DEA) records and expert educational materials, (2) real-time access to peer-reviewed scientific articles via external database connectivity, and (3) agentic architecture for reasoning over source selection, information synthesis, and conversational context management. The system enforces dual retrieval accessing locally stored information and external scientific repositories, supporting dynamic evidence-based response generation with credible source attribution.

The study examines whether dual-source retrieval architecture integrating local knowledge bases with real-time scientific literature queries provides more accurate and current responses than single sources. It evaluates the ability to maintain conversational coherence and contextual consistency across multi-turn interactions while retaining factual accuracy and transparent source attribution. Finally, it assesses system reliability, suitability, and usability in educational settings.

Thirty domain-specific questions measured system performance across substance use classifications, health effects, prevention methods, and regulatory policy. Context-dependent follow-up questions measured conversational memory and coherence. Five domain experts independently evaluated responses based on factual accuracy, citation relevance, contextual compatibility, and regulatory compliance. This study advances AI-mediated health education by demonstrating the efficacy of agentic architectures in providing scalable and transparent interventions. It provides an empirical model for integrating authoritative sources with conversational AI to improve substance use education delivery. These findings offer a template for developing reliable, evidence-based AI systems for health communication and behavior change.

## METHODOLOGY

This study employed a design science research approach encompassing the development, implementation, and expert evaluation of an agentic AI system for substance use education and prevention. The methodology comprised three phases: data curation, system architecture development, and expert assessment.

The knowledge base comprised 102 documents retrieved directly from official Drug Enforcement Administration (DEA) digital platforms. This included 71 publications sourced from the DEA's official website (drug fact sheets, policy manuals, and scheduling guidelines) and 31 transcriptions of expert-led content from the DEA's official YouTube channel, focusing on substance use prevention and policy. Selection was limited to materials that were publicly accessible, published or updated within the last five years, and primarily focused on education rather than clinical treatment protocols.

A multi-stage processing pipeline ensured high-fidelity data extraction from diverse sources. DEA publications in PDF format were processed via a tiered extraction framework: PyMuPDF4LLM [38] for primary text extraction, while OCRmyPDF [39] was programmatically triggered as a fallback for scanned documents where the Unicode replacement ratio exceeded 10%. For documents with complex structural elements such as tables and multi-column layouts, the Docling parser [40] was employed to preserve document hierarchy.

Video-based content was ingested using yt-dlp [41], with extraction logic prioritizing manually curated subtitles over auto-generated transcripts to ensure clinical and legal terminology accuracy. All data underwent standardized preprocessing including removal of HTML artifacts, whitespace normalization, and stripping of non-relevant metadata.

*Figure 1 illustrates the complete document processing pipeline from raw data to searchable vectors.*

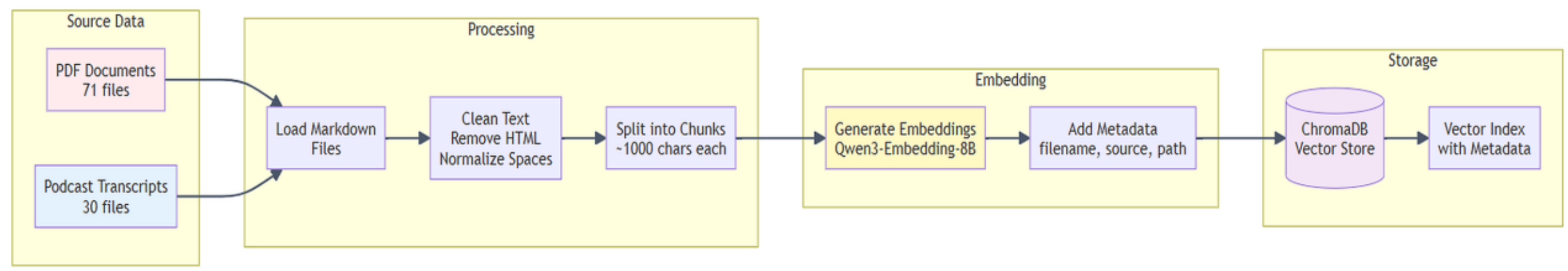

*Figure 1: Data processing pipeline showing the transformation of raw source materials into searchable vector representations.*

The extracted text was processed using paragraph-aware segmentation strategy to preserve semantic context. Chunks were targeted at approximately 1,000 characters, with the algorithm aggregating consecutive paragraphs while strictly adhering to paragraph boundaries. This approach aligns with empirical evidence suggesting that transformer-based embedding performance is optimized when context windows are maintained between 256 and 1,024 tokens [42].

These segments were transformed into 1,024-dimensional dense vector representations using the Qwen3-Embedding-8B model [43]. The resulting embeddings were persisted in a ChromaDB vector database [44]. Retrieval efficiency was optimized through Hierarchical Navigable Small World (HNSW) indexing, allowing for scalable and rapid nearest-neighbor queries within the high-dimensional latent space [45].

The platform is built upon a modular five-layer architecture designed for high-availability and verifiable health communication: (1) Presentation Layer (Streamlit [46]); (2) Orchestration Layer (LangChain-based agent coordination [47]); (3) Retrieval Layer (Dual-source knowledge access); (4) Generation Layer (Qwen3-32B large language model); and (5) Storage Layer (ChromaDB for vector persistence and persistent session management).

*Figure 2 provides an overview of the complete system architecture and component interactions.*

*Figure 2: System architecture overview illustrating the five-layer design. The presentation layer (Streamlit) handles user interaction; the application layer (LangChain agent with RAG chain and*

*context-aware retriever) orchestrates reasoning and tool coordination; the data processing layer manages text cleaning, splitting, and embedding generation; the storage layer (ChromaDB vector database (ChromaDB, [44]) implemented through LangChain framework [47] with HNSW indexing) maintains document vectors and chat history; and external services (Qwen3-32B LLM and Qwen3-Embedding-8B via API) provide language generation and semantic encoding capabilities.*

.

The Retrieval Layer employs a hybrid mechanism synthesizing information from two distinct domains. The primary mechanism performs local vector similarity searches within ChromaDB, retrieving the top three (k=3) semantically relevant document segments from the DEA corpus. Simultaneously, the system interfaces with the National Center for Biotechnology Information (NCBI) E-utilities API [48], allowing the agent to programmatically query PubMed for contemporary peer-reviewed literature, specifically prioritizing systematic reviews and high-impact publications from the last five years.

To ensure information integrity, the system implements contextual source weighting. Orchestration logic provides the model with explicit heuristics: DEA records are assigned higher weight for queries involving legal scheduling, regulatory compliance, and federal policy, whereas PubMed-derived data is prioritized for inquiries regarding neurobiological mechanisms, pharmacological efficacy, and clinical outcomes.

*Figure 3 illustrates the system's user interface, designed to prioritize transparency and accessibility for educational users.*

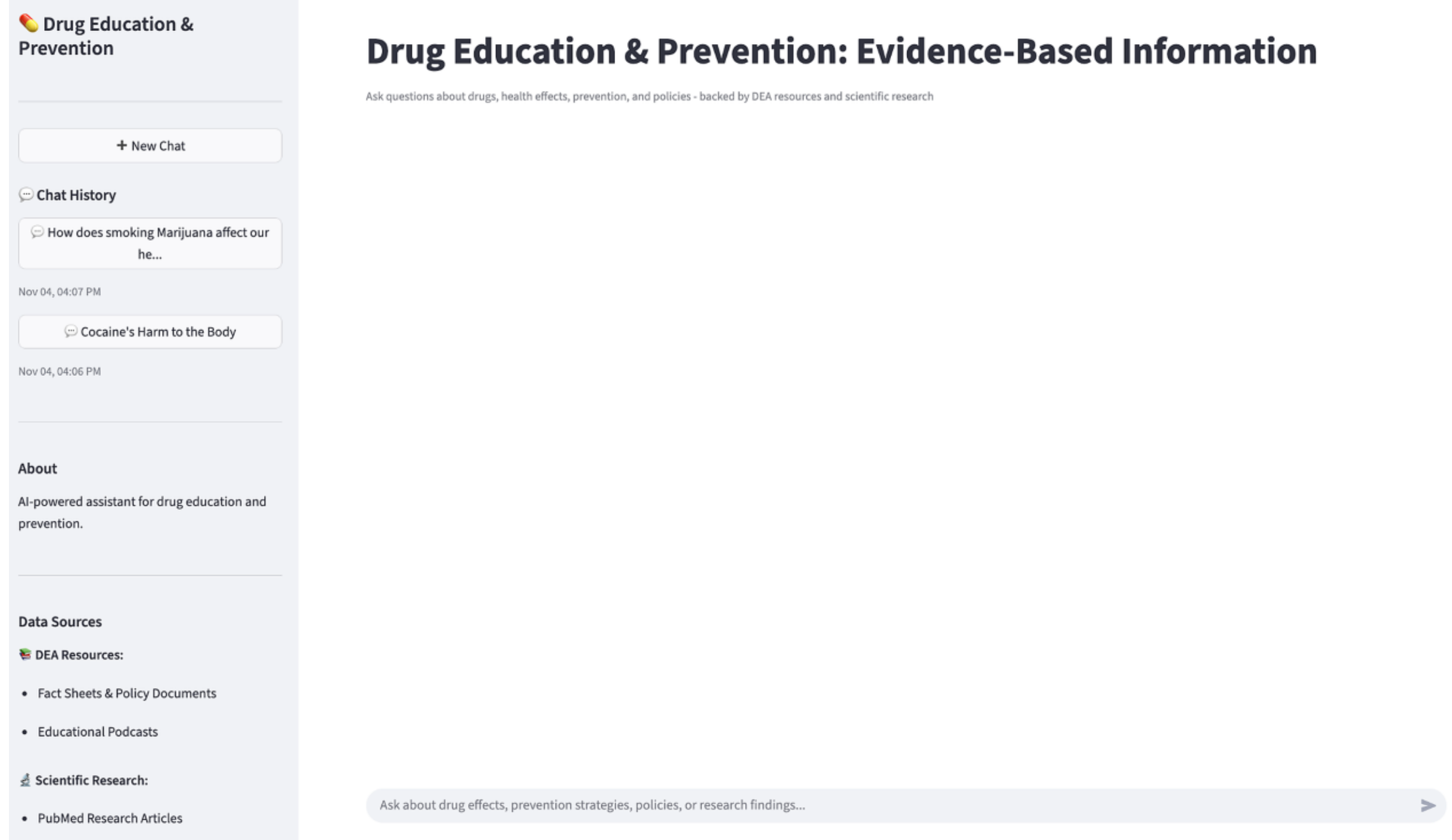

*Figure 3: System user interface showing conversational input field, data source sidebar, and chat history management. The interface displays accessible DEA resources and PubMed scientific literature sources, with visible conversation history enabling users to track multi-turn interactions.*

External tool communication was facilitated through the Model Context Protocol (MCP), a standardized framework for enabling secure, structured interaction between language models and external data ecosystems. By utilizing JSON-RPC 2.0 messaging [49], MCP allowed for decoupled and modular integration of the PubMed API. Although MCP is a nascent protocol, its implementation serves as a proof-of-concept for creating interoperable AI systems that can seamlessly bridge the gap between static local knowledge bases and dynamic global research repositories.

The orchestration layer maintains conversational state across interactions, dynamically selects appropriate tools, and coordinates multi-step reasoning processes. For each query, a history-aware retriever reformulates questions, incorporating previous dialogue context, transforming follow-ups like "How does it compare to heroin?" into "How do the health effects of fentanyl compare to heroin?" when fentanyl was discussed previously.

Generation operates through structured prompts specifying that responses must be grounded in retrieved documents, include supporting citations, acknowledge uncertainty when information is insufficient, and maintain an educational tone for general audiences. Conversational memory persists across sessions, enabling users to resume dialogues and developers to analyze interaction patterns.

*Figure 4 illustrates the system's internal reasoning process during query processing.*

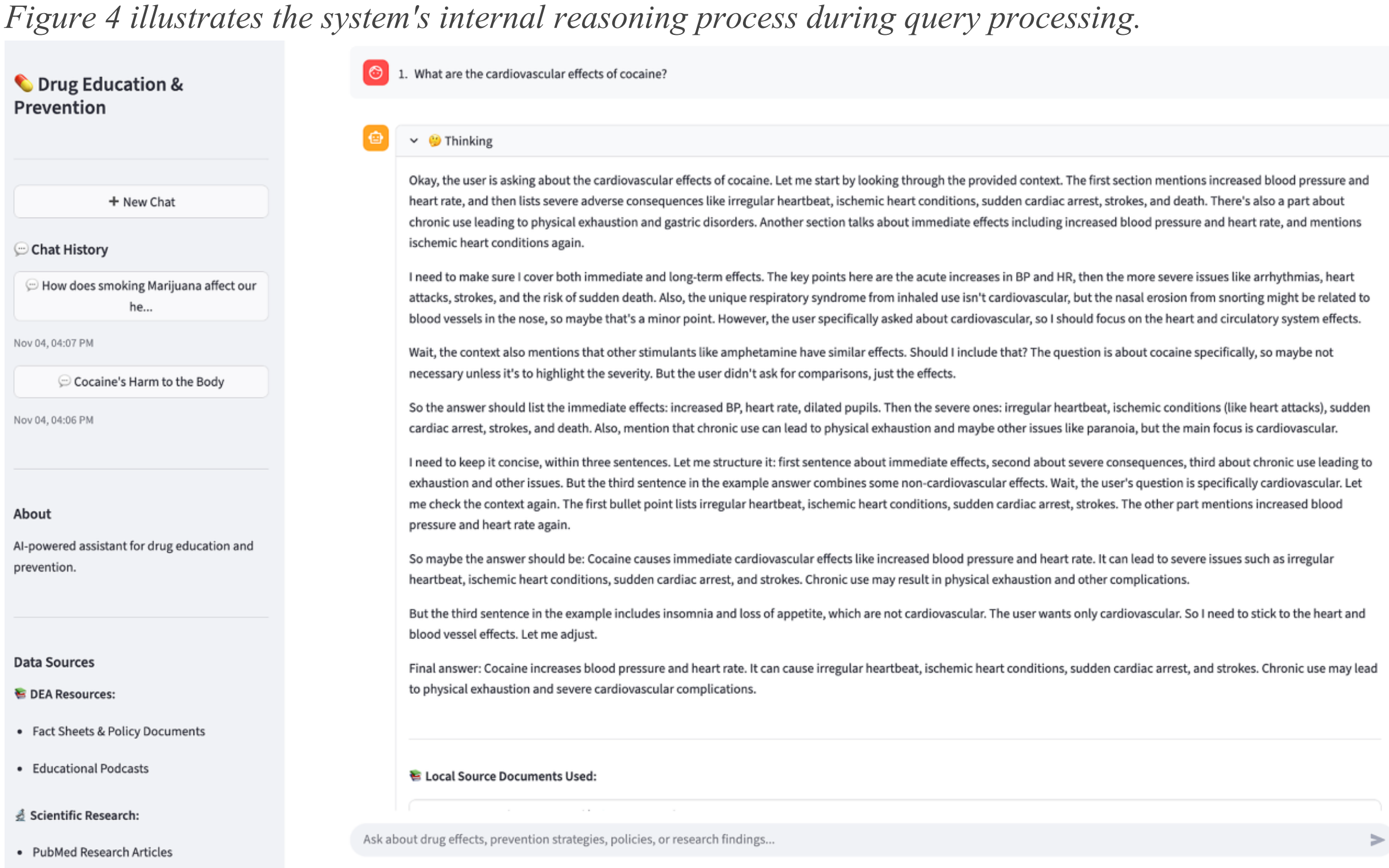

*Figure 4: The system reasoning process illustrates how the agentic orchestration layer analyzes query requirements, evaluates retrieved document context, assesses relevance to the user's question, and*

*determines the optimal response structure before generating the final output. This transparency enabled evaluators to understand decision-making logic and identify areas for refinement*

We recruited a panel of five independent subject matter experts to facilitate both generation and evaluation of educational content. The experts included professionals with advanced degrees in public health, clinical social work, and addiction medicine, with an average of 14 years of experience in substance use prevention and policy.

The evaluation was conducted in three phases. In the first phase, experts generated 50 unique questions based on their professional experience and core domains covered by DEA and PubMed literature (drug scheduling, neurobiological effects, and prevention policy). In the second phase, we applied semantic deduplication using Sentence-BERT (SBERT) [50] to generate vector embeddings for all 50 questions and calculated cosine similarity scores; questions exhibiting semantic similarity of 90% or higher were considered redundant and removed. This resulted in a final set of 30 distinct, high-priority questions.

The five experts collaboratively posed these 30 questions to the system, along with two contextually dependent follow-up queries for each, totaling 90 system interactions, to assess conversational memory and coherence. To ensure independent evaluation, two members of the expert panel were selected as formal raters and assessed all 90 interactions without knowledge of each other's ratings. Responses were rated on a five-point Likert scale across four criteria: (1) Factual Accuracy; (2) Citation Quality; (3) Contextual Coherence; and (4) Regulatory Appropriateness, where 1 = Poor, 2 = Below Adequate, 3 = Adequate, 4 = Good, and 5 = Excellent. Inter-rater reliability was assessed using Cohen's kappa (κ) to quantify agreement between the two raters.

This study was conducted in accordance with ethical standards for medical informatics research. The panel of five subject matter experts participated anonymously. The research utilized exclusively publicly available documents from the DEA and PubMed; therefore, it was exempt from Institutional Review Board oversight. The platform provides explicit source attribution for every response and includes a disclaimer stating that the tool is for educational purposes and does not substitute for professional medical advice.

Large Language Model Use: The Qwen3-32B large language model served as the generative component of the system described in this study and is integral to the architecture evaluated herein. AI-assisted tools were additionally used for copyediting purposes during manuscript preparation, consistent with Springer Nature’s definition of AI-assisted copy editing. No AI tool was used for autonomous content creation, data analysis, or interpretation of results.

## RESULTS

Independent evaluation by two subject matter experts analyzed 90 system interactions using four assessment criteria: factual accuracy, citation quality, contextual coherence, and regulatory appropriateness. The system demonstrated effective synthesis of diverse information sources while maintaining conversational coherence and producing educationally appropriate responses. Expert evaluations yielded consistently high ratings across all criteria. Mean scores on the five-point scale were 4.31 (SD=0.68) for factual accuracy, 4.18 (SD=0.74) for citation quality, 4.26 (SD=0.71) for contextual coherence, and 4.35 (SD=0.63) for regulatory appropriateness, indicating performance predominantly in the "Good" to "Excellent" range.

**Table 1: Expert Evaluation Scores Across Four Assessment Criteria**

| Evaluation Criterion | Mean Score (SD) | Range | Performance Level |
|---|---|---|---|
| Factual Accuracy | 4.31 (0.68) | 3–5 | Good–Excellent |
| Citation Quality | 4.18 (0.74) | 3–5 | Good–Excellent |
| Contextual Coherence | 4.26 (0.71) | 3–5 | Good–Excellent |
| Regulatory Appropriateness | 4.35 (0.63) | 3–5 | Good–Excellent |
| **Overall Mean** | **4.28 (0.69)** | **3–5** | **Good–Excellent** |

*Table 1: Expert evaluation scores across four assessment criteria (N = 90 system interactions, two independent raters, five-point Likert scale). SD = standard deviation.*

Inter-rater reliability assessed using Cohen's kappa for categorical judgments (acceptable vs. unacceptable at rating $\geq 3$) was $\kappa = 0.78$, indicating substantial agreement. No responses received ratings below "Adequate" from both experts simultaneously, suggesting the system avoided catastrophic failures or harmful misinformation generation. Performance varied by question category. Substance use classifications and scheduling (mean: 4.52) and regulatory policy (mean: 4.48) received the highest scores, reflecting effective DEA documentation retrieval. Health effects and risks (mean: 4.21) and prevention strategies (mean: 4.19) performed well but showed greater variability due to multi-source synthesis requirements. Treatment and recovery resources (mean: 4.05) received the lowest ratings, primarily because local DEA documents contained less treatment information compared to prevention topics, requiring greater PubMed reliance with attendant challenges of abstract-level information and technical terminology.

**Table 2: System Performance by Question Category**

| Question Category | Mean Score | Primary Knowledge Source |
|---|---|---|
| **Drug Classifications & Scheduling** | 4.52 | DEA Documentation |
| **Regulatory Policy** | 4.48 | DEA Documentation |
| **Health Effects & Risks** | 4.21 | Dual-source (DEA + PubMed) |
| **Prevention Strategies** | 4.19 | Dual-source (DEA + PubMed) |

| **Treatment & Recovery Resources** | 4.05 | PubMed (primary) |
|---|---|---|

*Table 2: System performance by question category showing mean expert ratings and primary knowledge source utilized (N = 30 questions).*

Analysis of retrieval logs revealed distinct patterns. The system retrieved relevant local documents for 87% of queries. PubMed queries returned relevant abstracts for 73% of questions, with lower success for regulatory topics. For questions where both retrieval mechanisms succeeded, experts noted effective integration of regulatory guidance with scientific evidence. One expert noted the response balanced legal and clinical evidence effectively. Questions with lower retrieval quality in one or both sources produced less comprehensive responses. Questions about emerging substances not yet widely studied (e.g., novel synthetic opioids) relied primarily on DEA intelligence reports, providing classification and enforcement information but limited pharmacological detail. These responses were factually accurate but less educationally complete than those benefiting from dual-source retrieval.

The present study does not include a controlled comparative baseline, which is an acknowledged limitation discussed in the Limitations section below.

*Figure 5 demonstrates this dual-source integration capability with a representative example query and response.*

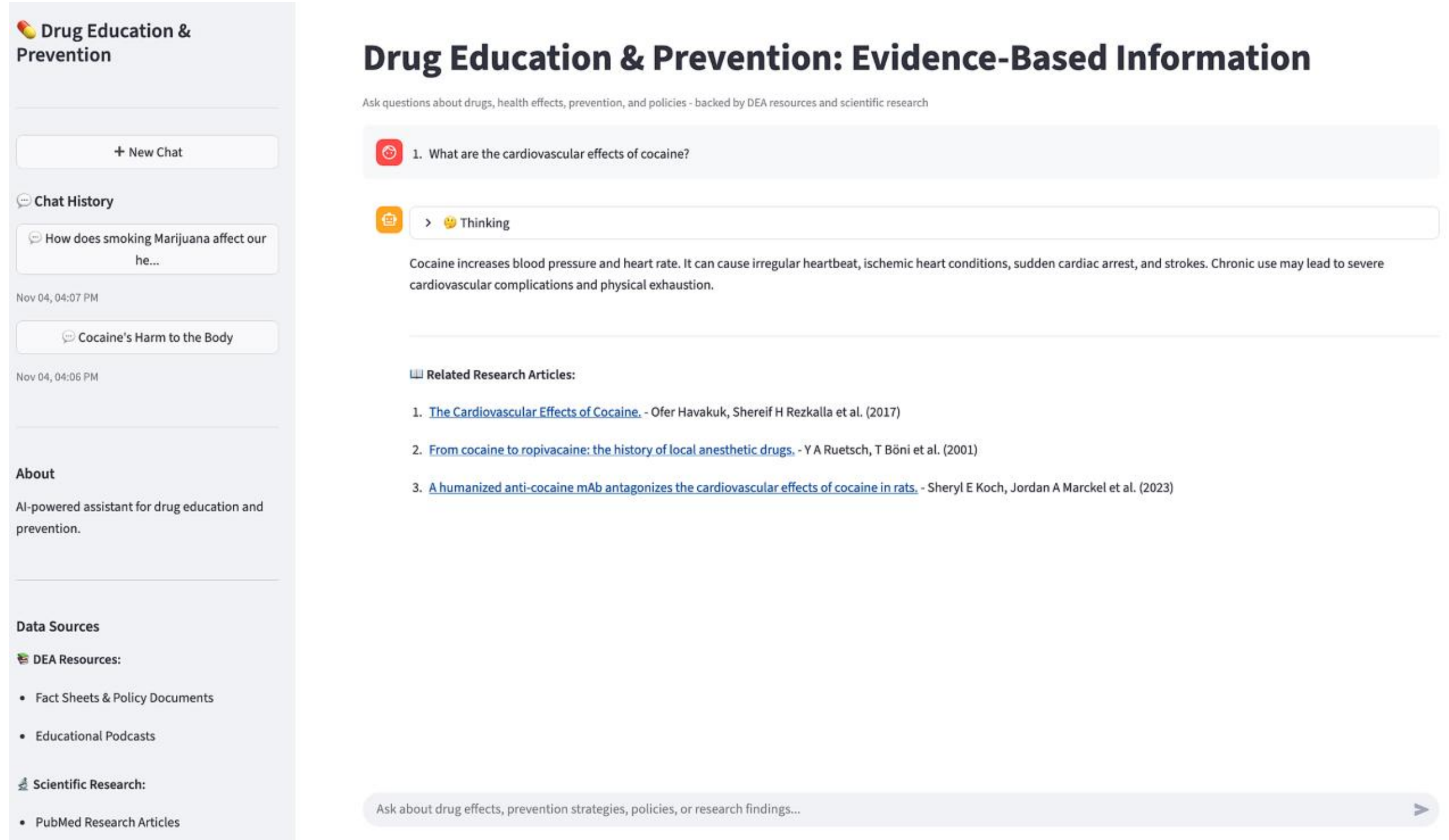

*Figure 5: Example system response to a query about cardiovascular effects of cocaine, demonstrating dual-source retrieval and transparent citation. The response synthesizes information from DEA*

*documentation and peer-reviewed PubMed literature, with each claim supported by visible source attribution, including document titles and relevance scores.*

Source weighting functioned appropriately in most cases. DEA documents were cited for regulatory questions, while peer-reviewed articles were emphasized for health-related queries. However, in 12% of responses, experts noted citation balance could improve, suggesting opportunities for refining retrieval ranking. The system maintained context across multi-turn interactions effectively. Of 60 follow-up queries, 93% were interpreted correctly with appropriate incorporation of previous dialogue context. This high success rate demonstrates effective conversational memory and history-aware retrieval implementation.

Representative examples illustrate this capability. After asking "What schedule is methamphetamine?" a follow-up "Why is it classified there instead of Schedule I?" correctly interpreted "there" as Schedule II, accepted medical use. Context interpretation failures occurred in 7% of follow-up queries, typically involving ambiguous pronouns with multiple possible referents. After discussing both prescription opioids and heroin, "How common is its use among adolescents?" generated a response about prescription opioids when the user likely intended heroin. These failures highlight robust anaphora resolution challenges and suggest potential value for clarification mechanisms when context is ambiguous.

*Figure 6 shows the source attribution interface that evaluators found particularly valuable for verification.*

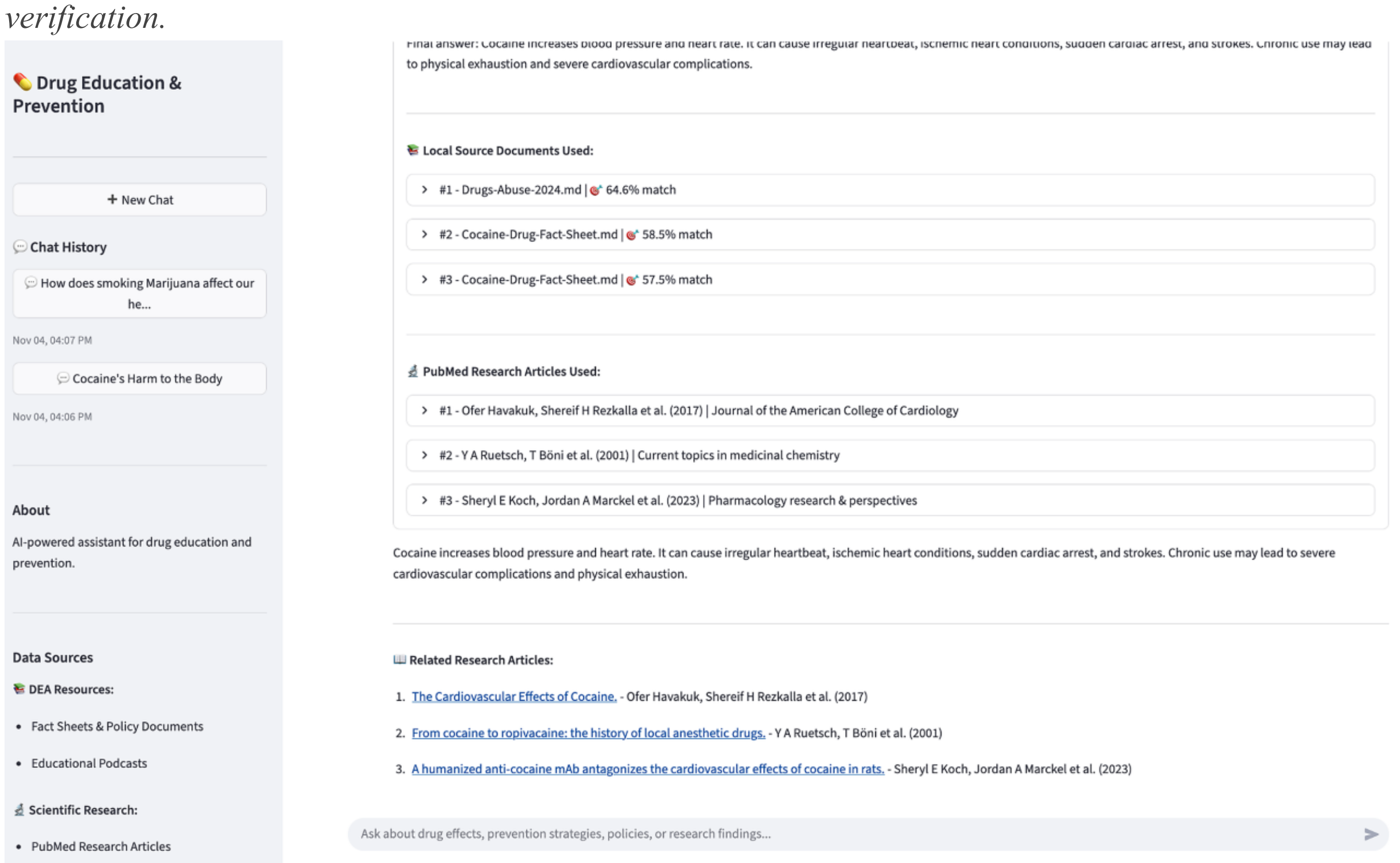

*Figure 6: Source attribution interface displaying both local DEA documents with relevance match percentages (e.g., "Cocaine-Drug-Fact-Sheet.md | 58.5% match") and PubMed research articles with full*

*bibliographic details, author information, publication years, and direct hyperlinks. This dual-source transparency enables independent verification of all factual claims.*

References were most suitable when connecting definite facts to reliable sources. A reply that fentanyl is about 50 times stronger than heroin referenced a particular DEA fact sheet to make the comparison. Citations were deemed inappropriate when referencing sources with useful information but requiring users to make inferences, or when multiple references were made to the same claim when only one was needed.

Transparency was appreciated by experts. Both observed that attributing visible sources enhanced confidence and verification. One expert remarked that it was easy to see the specific DEA document or research article that supported every argument. This transparency would be key to trust building as there are valid concerns about AI-generated misinformation. Citation details, however, had readability tradeoffs. Responses that included many references were sometimes cluttered with different sources making similar points. Experts indicated that approximately 15% of responses could benefit from consolidated or simplified citations while preserving verifiability.

Qualitative expert feedback highlighted several strengths. Responses were consistently clear, well-structured, and generally suitable for non-specialists. The system appropriately balanced technical and accessible language, used logical structure defining important terms, providing main information, and giving context or implications. The system was cautious in responding to sensitive issues. Regarding questions about drug effects or risks, answers acknowledged individual and situational differences without overgeneralization. Instead of presenting information that could facilitate harm, like comprehensive synthesis instructions, the system guided users to sources of prevention, education, and treatment.

Nevertheless, response quality varied. Approximately 18% of answers were deemed more technical than necessary for general audiences, generally when synthesizing information from scientific abstracts using specialized language. One answer regarding opioid mechanisms contained phrases like "mu-opioid receptor agonism" without adequate explanation for non-experts, suggesting need for further refinement to ensure accessibility. Hedging language occasionally appeared in responses, phrases such as "research suggests" or "evidence indicates" which, while appropriate for expressing uncertainty, were occasionally repetitive or disruptive of clarity. Balancing epistemic humility with confident communication of facts remains a design challenge requiring further refinement. The 18% technicality rate carries direct design implications: future iterations should incorporate adaptive verbosity mechanisms that detect user literacy level, through explicit profile settings or implicit signals such as query complexity, and adjust response language accordingly. This aligns with established health literacy principles recommending that patient-facing health information target a general reading level [36], and represents a tractable engineering improvement that could substantially broaden the system's accessible user base.

Analysis of lower-rated responses revealed systematic limitations. The most common failure mode involved insufficient retrieval when queries were vague, used non-standard terminology, or addressed underrepresented topics. A question about "club drugs" initially retrieved MDMA documents but missed relevant GHB and ketamine materials because the query term did not appear prominently in those documents. Query reformulation or expansion strategies might address this limitation.

A second failure mode involved over-reliance on abstract-level PubMed information when full-text articles would provide necessary context. Abstracts sometimes presented research findings without sufficient methodological detail or nuance, leading to responses stating results without adequately qualifying them. While full-text access would improve this situation, licensing and technical constraints limited retrieval to abstracts.

## DISCUSSION

This study demonstrates that agentic AI architectures integrating authoritative knowledge sources with real-time literature retrieval can provide accurate, transparent, and educationally appropriate substance use education support. Results indicate performance levels sufficient for practical deployment in educational contexts.

The consistently high expert ratings across evaluation criteria indicate that retrieval-augmented generation offers a robust solution to knowledge currency and verifiability challenges that plague standalone language models. By grounding responses in explicitly retrieved authoritative sources, the system circumvented fabrication errors common in purely generative approaches. Notably, expert raters did not identify any responses as factually incorrect within the scope of this evaluation, a promising finding, though one that should be interpreted cautiously given the limited sample of 30 questions and two raters. Larger-scale evaluation would be necessary to make stronger claims about error rates, particularly for novel or edge-case queries.

Performance variation across question categories provides insights into the relationship between knowledge base composition and system capability. Questions concerning drug classifications and regulatory policy, where DEA documentation is comprehensive and authoritative, achieved the highest scores. Questions requiring synthesis of pharmacological mechanisms or treatment approaches showed greater variability. This pattern suggests that system performance depends not only on retrieval architecture but on the availability and quality of relevant sources in the knowledge base.

The successful integration of both local and external retrieval represents an important architectural contribution. Previous AI-based education systems typically employ either curated local knowledge or external databases, but not both. Our findings demonstrate that dual-source retrieval can achieve a balance between regulatory accuracy and scientific currency that single-source approaches cannot match. This observation extends beyond substance use education, suggesting broader applicability to domains requiring integration of authoritative policy guidance with evolving scientific evidence, such as environmental health, occupational safety, or public health emergency response.

The system's capacity to maintain context across multi-turn interactions distinguishes it from conventional information retrieval systems and represents meaningful progress toward conversational AI supporting natural learning progressions. Educational research consistently demonstrates that learning proceeds through iterative questioning [29]. Conversational memory reduces cognitive burden by enabling follow-up questions that build naturally on previous exchanges. Expert feedback specifically highlighted this capability as enhancing educational utility. The observation that conversational continuity "would encourage learners to ask deeper questions" points to a potentially significant benefit: not merely answering questions users already have but facilitating the question-asking process itself. However, validating this hypothesis requires longitudinal user studies measuring learning outcomes, engagement patterns, and knowledge retention, evaluation dimensions beyond the present study's scope but critical for future research.

The 7% context interpretation failure rate highlights persistent challenges in robust natural language understanding. These failures suggest complementary value in clarification mechanisms enabling systems to ask users for disambiguation when context is uncertain, balancing fluid conversation goals against excessive interruption risks. The system's transparent source citation represents both a technical feature and an ethical commitment to user agency and informed decision-making. Expert evaluators emphasized that visible source attribution increased confidence and facilitated verification, consistent with broader research demonstrating that transparency about system reasoning significantly influences user acceptance [28,29].

Findings suggest that retrieval-augmented generation can provide scalable methods for delivering accurate, source-grounded information. From a policy perspective, findings point to AI's potential to strengthen public health education infrastructure while highlighting governance challenges. Access to authoritative government documentation and scientific literature was critical to system performance.

However, citation transparency involves design tradeoffs. Evaluator feedback noting some responses felt "cluttered" with citations suggests the need for adaptive approaches, adjusting citation detail based on user sophistication or query context. The finding that citations were most valuable when linking specific claims to specific sources has implications for prompt engineering and response generation strategies.

## LIMITATIONS

Several constraints limit the generalizability and interpretation of findings. The evaluation employed 30 questions and 90 total interactions rated by two independent experts from a panel of five, providing valuable initial evidence but limited statistical power. Experts assessed accuracy, appropriateness, and coherence, but did not measure whether accurate responses actually improve user knowledge, change attitudes or intentions, or influence behavior, the ultimate outcomes of educational interventions. Establishing educational impact would require longitudinal studies assessing knowledge gains, retention, and application among actual learners.

The present study does not include a controlled comparative baseline. A rigorous evaluation of the dual-source architecture would require at a minimum three ablation conditions: (1) a DEA-only retrieval condition; (2) a PubMed-only condition; and (3) a general-purpose LLM condition without retrieval augmentation. While such a formal ablation was not conducted, indirect evidence supports the value of the dual-source approach. Giorgi et al. (2024) [35] evaluated general-purpose LLMs responding to real-world drug-related questions without retrieval augmentation and documented substantial accuracy limitations. Similarly, Lee et al. [15] found that standalone chatbot interventions for substance use prevention operated without authoritative knowledge integration, limiting factual reliability. The performance gradient observed in the present study, with DEA-rich regulatory topics scoring highest (mean: 4.52) and PubMed-dependent treatment topics scoring lowest (mean: 4.05), provides indirect evidence that knowledge source availability directly moderates response quality, consistent with what would be predicted if either source were removed. Future work should empirically test these hypotheses through a controlled comparative study.

The knowledge base reflects the U.S. regulatory framework and may not generalize to international contexts where substance use policies and educational priorities differ. The system's focus on educational applications intentionally excludes clinical decision support, limiting broader utility. Current results represent Kirkpatrick Level 2 evaluation (learning and knowledge accuracy) but leave Level 3 (behavioral transfer) unaddressed [51,52]. Future research must transition to user-centered longitudinal studies measuring objective knowledge gain, self-efficacy [53], and changes in health-seeking intentions among actual learners [15]. Rigorous experimental designs such as randomized controlled trials are necessary to determine whether engagement translates into real-world behavioral outcomes [54].

## CONCLUSION

Substance use education operates in a high-stakes environment where misinformation carries real consequences for individual and public health. This study demonstrates that agentic AI architectures integrating authoritative regulatory documentation with real-time peer-reviewed literature can meet the accuracy, transparency, and contextual consistency demands of this domain. Expert evaluation across 90 system interactions yielded strong performance on all four criteria, factual accuracy (mean: 4.31), citation quality (mean: 4.18), contextual coherence (mean: 4.26), and regulatory appropriateness (mean: 4.35) on a five-point scale, with high inter-rater agreement ($\kappa$=0.78) and no responses rated below "Adequate," indicating the system avoids the catastrophic misinformation failures that disqualify general-purpose LLMs from health education contexts.

Beyond the technical results, this work advances a replicable design principle: that grounding conversational AI in dual authoritative sources, domain-specific regulatory corpora and dynamic scientific literature, can resolve the core limitations of parametric language models in knowledge-intensive educational settings. The architecture's transparency mechanisms, including visible source attribution and relevance scoring, directly address the "black box" problem that undermines educator and learner trust in AI-generated health information.

These findings carry practical implications for health education developers, public health agencies, and policymakers. The system architecture is domain-transferable and could be adapted for other fields requiring real-time synthesis of regulatory guidance and scientific evidence, including environmental health, occupational safety, and emergency public health response. While validation through longitudinal user studies and randomized controlled trials remains necessary to establish behavioral impact, the present results provide a rigorous foundation and a reproducible framework for evidence-based AI health education at scale.

## DECLARATIONS

**AI Use Disclosure:** The Qwen3-32B large language model is the generative component of the system described and evaluated in this study. AI-assisted tools were additionally used for copyediting during manuscript preparation (grammar, spelling), consistent with Springer Nature's definition of AI-assisted copy editing. No AI tool was used for autonomous content creation, data analysis, or interpretation of results.

**Author Contributions:** KH conceptualized the study, designed and implemented the agentic AI system and RAG pipeline, conducted the expert evaluation, performed data analysis, and wrote the original manuscript. ZK contributed to system architecture design, knowledge base curation, and critical revision of the manuscript. MA supervised the research, provided methodological guidance, and reviewed and approved the final manuscript. All authors read and approved the submitted version.

**Funding**: This research received no specific grant from any funding agency.

**Competing Interests**: The authors declare no competing interests.

**Ethics Approval**: Ethical review and approval were waived for this study because it utilized a publicly available dataset, where all data were de-identified.

**Informed Consent Statement**: Not applicable. All data used in the study were de-identified.

**Data Availability:** The 30 evaluation questions, rating rubric, and anonymized inter-rater scoring data are available from the corresponding author upon reasonable request. The system codebase, including the RAG pipeline, retrieval configuration, and prompt templates, will be made publicly available via a GitHub repository upon acceptance. DEA source documents are publicly available at https://www.dea.gov/factsheets. PubMed data were retrieved via the NCBI E-utilities API (https://www.ncbi.nlm.nih.gov/books/NBK25497/).

**Clinical Trial Number**: Not applicable.